\relax
\documentclass[letterpaper]{article} 
\usepackage{aaai18}  
\usepackage{times}  
\usepackage{helvet}  
\usepackage{courier}  
\usepackage{url}  
\usepackage{graphicx}  
\usepackage{amsmath}
\usepackage{amsfonts}
\usepackage{amssymb}
\usepackage{amsthm}
\usepackage{subfigure}
\usepackage{graphicx}
\newtheorem{corollary}{Corollary}

\newtheorem{theorem}{Theorem}
\usepackage{algorithm}
\usepackage[noend]{algpseudocode}
\usepackage{booktabs}       

\begin{document}
%
\title{Efficient Probabilistic Performance Bounds for Inverse Reinforcement Learning}
\author{ Daniel S. Brown \and   Scott Niekum \\
 Department of Computer Science \\
  University of Texas at Austin \\
  \texttt{\{dsbrown,sniekum\}@cs.utexas.edu} \\
}
\maketitle
\begin{abstract}
In the field of reinforcement learning there has been recent progress towards safety and high-confidence bounds on policy performance. However, to our knowledge, no practical methods exist for determining high-confidence policy performance bounds in the inverse reinforcement learning setting---where the true reward function is unknown and only samples of expert behavior are given. We propose a sampling method based on Bayesian inverse reinforcement learning that uses demonstrations to determine practical high-confidence upper bounds on the $\alpha$-worst-case difference in expected return between any evaluation policy and the optimal policy under the expert's unknown reward function. We evaluate our proposed bound on both a standard grid navigation task and a simulated driving task and achieve tighter and more accurate bounds than a feature count-based baseline. We also give examples of how our proposed bound can be utilized to perform risk-aware policy selection and risk-aware policy improvement. Because our proposed bound requires several orders of magnitude fewer demonstrations than existing high-confidence bounds, it is the first practical method that allows agents that learn from demonstration to express confidence in the quality of their learned policy.
\end{abstract}

\section{Introduction}
There is a growing interest in safety and risk-sensitive metrics for machine learning and artificial intelligence systems, especially for systems that interact with their environment \cite{garcia2015comprehensive,amodei2016concrete,thomas2017ensuring}. 
Risk-aware approaches have been recently proposed and applied to many different problems including planning in Markov decision processes \cite{chow2015risk}, physical search problems \cite{brown2016exact}, reinforcement learning \cite{tamar2015optimizing,garcia2015comprehensive}, and imitation learning \cite{santara2017rail}; however, to the best of our knowledge, no one has investigated how to obtain sample-efficient, risk-aware confidence bounds on the performance of a policy under an unknown reward function, as is the case when learning from demonstrations. 

Learning from demonstration (LfD) is a popular method to learn a skill or policy by simply observing demonstrations from an expert \cite{Argall2009}. 
One popular variant of LfD is Inverse Reinforcement Learning (IRL) \cite{ng2000algorithms} where the goal is to infer a reward function that explains the demonstrated behavior.
LfD techniques based on IRL have potential applications in many settings such as manufacturing, home and hospital care, and autonomous driving. In these types of real-world settings it is important, and perhaps critical, to provide performance bounds on an agent's learned policy. For example, consider a hospital assistant robot that has learned from demonstrations how to lift a patient out of bed. Before deploying this learned policy, we would want to provide a high-confidence bound on the difference in performance between the robot's learned policy and the optimal policy under the expert's reward. 
If this bound on policy loss is too high, then the robot could request additional demonstrations until, with high confidence, its policy loss with respect to the optimal policy is within some allowable error margin.

We propose a general method for obtaining high-confidence performance bounds in the inverse reinforcement learning setting---where the true reward function is unknown and only samples of expert behavior are given. 
Given demonstrated trajectories of a task, our goal is to allow an agent to bound the difference in expected return between the agent's own policy and the optimal policy for the task, under the expert's unknown reward function. Because the problem of Inverse Reinforcement Learning is ill-posed (there are an infinite number of reward functions that result in the same optimal behavior), we seek a risk-sensitive bound on this performance difference that takes into account the uncertainty in the posterior distribution over reward functions, conditioned on the demonstrations.




We perform Markov chain Monte Carlo sampling using Bayesian Inverse Reinforcement Learning \cite{ramachandran2007bayesian} to sample likely reward functions given the demonstrations. Using these sampled reward functions, we compute samples of the expected return difference between the optimal policy under the expert's reward function and the agent's policy. These samples are then used to calculate a $(1-\delta)$ probabilistic upper bound on the $\alpha$-worst-case policy loss. We obtain this bound without knowing the expert's policy or true reward function.

Our main contributions are
(1) we formalize the problem of high-confidence policy evaluation in the inverse reinforcement learning domain; (2) we present the first practical method for obtaining high-confidence bounds on the $\alpha$-worst-case difference in expected return between any evaluation policy and the optimal policy under the expert's unknown reward function; (3) we evaluate our proposed bound on standard grid navigation and simulated driving tasks and demonstrate a significant improvement over existing statistical bounds and an empirical baseline based on feature counts; and (4) we give examples of how our proposed bound enables risk-aware policy ranking and risk-aware policy improvement, given only demonstrations of a task.



%


\section{Preliminaries} \label{sec:Preliminaries}
\subsection{Markov decision processes}
A Markov decision process (MDP) is defined as a tuple $\langle S, A, T, R, \gamma, S_0 \rangle$ where $S$ is the set of states, $A$ is the set of actions, $T:S \times A \times S \to [0,1]$ is the transition function, $R: S \to \mathbb{R}$ is the reward function, $\gamma \in [0,1)$ is the discount factor, and $S_0$ is the initial state distribution. 

A policy $\pi$ is a mapping from states to a probability distribution over actions. The value of a policy $\pi$ under reward function $R$ is the expected return of that policy and is denoted as $V^\pi_R = \mathbb{E}_{s_0 \sim S_0}[\sum_{t=0}^\infty \gamma^t R(s_t) | \pi]$. The value of executing policy $\pi$ starting at state $s \in S$ is defined as 
$V^\pi_R(s) = \mathbb{E}[\sum_{t=0}^\infty \gamma^t R(s_t) | \pi, s_0 = s]$. 
Given a reward function $R$, the Q-value of a state-action pair $(s,a)$ is defined as $Q^{\pi}_R(s,a) = R(s) + \gamma \sum_{s' \in S} T(s,a,s') V^\pi_R(s')$. We denote $V^*_R = \max_{\pi} V^\pi_R$ and $Q^*_R(s,a) = \max_{\pi} Q^{\pi}_R(s,a)$.

As is common in the literature \cite{abbeel2004apprenticeship,ziebart2008maximum}, we assume that the reward function can be expressed as a linear combination of features, so that $R(s) = w^T \phi(s)$ where $w \in \mathbb{R}^k$ is the k-dimensional vector of feature weights. Thus, we can write the value of a policy as  
$V^\pi_R \,= \,\mathbb{E}_{s_0 \sim S_0}[\sum_{t=0}^{\infty} \gamma^t w^T \phi(s_t) | \pi]
\,=\, w^T \mu(\pi)$,
where 
$\mu(\pi) = \mathbb{E}_{s_0 \sim S_0}[\sum_{t=0}^{\infty}\gamma^t \phi(s_t) | \pi]$
are the expected feature counts. Note that this does not affect the expressiveness of the reward function since $\phi$ can be a non-linear function. Given $\phi$, the reward function is fully specified by the feature weights $w$. Thus, we refer to the feature weights $w$ and the reward function $R$ interchangeably.

\subsection{Bayesian inverse reinforcement learning} \label{subsec:BIRL}
In IRL we are given an MDP without a reward function, denoted MDP$\setminus$R. Given a set of demonstrations, $D = \{(s_1,a_1),\ldots,(s_m,a_m)  \}$, consisting of state-action pairs, the IRL problem is to recover the reward function, $R^*$, of the demonstrator. Because this problem is ill-posed, IRL algorithms use a variety of heuristics and simplifying assumptions to find an estimate of $R^*$ \cite{gao2012survey}.

Bayesian IRL (BIRL) \cite{ramachandran2007bayesian} seeks to estimate the posterior over reward functions given demonstrations, $P(R|D)\propto P(D | R) P(R)$.
BIRL makes the assumption that the demonstrator is following a softmax policy, resulting in the likelihood function 
\begin{equation} \label{eqn:birl_likelihood}
P(D | R) = \prod_{(s,a) \in D} P((s,a)|R) = \prod_{(s,a) \in D}\frac{e^{c Q^*_R(s,a)}}{ \sum_{b \in A} e^{c Q^*_R(s,b)}}
\end{equation}
where $Q^*_R(s, a)$ is the optimal Q-value function for reward $R$, and $c$ is a parameter representing the confidence in the demonstrator's optimality. Equation \ref{eqn:birl_likelihood} gives greater likelihood to rewards for which the actions taken by the expert have higher Q-values than the alternative actions.


The softmax distribution over actions is commonly used as a likelihood function in IRL \cite{babes2011apprenticeship,levine2011nonlinear,michini2012bayesian,rothkopf2013modular} and has been empirically shown to be an effective model of human behavior, enabling accurate learning from human demonstrations \cite{lopes2007affordance,kim2016socially} and prediction of human actions \cite{baker2009action,karasev2016intent}.

The BIRL algorithm uses Markov chain Monte Carlo (MCMC) sampling to sample from the posterior $P(R|D)$. 
Feature weights are sampled according to a proposal distribution, and for each sample the MDP is solved to obtain the sample's likelihood and determine the transition probabilities within the Markov chain. For each new sample, the resulting MDP can typically be quickly solved by starting with the policy from the previous MDP and using only a few steps of policy iteration \cite{ramachandran2007bayesian}. 
An estimate of the expert's reward function can be found by averaging the feature weights in the chain to obtain the mean reward function \cite{ramachandran2007bayesian} or by using the maximum a posteriori (MAP) estimate \cite{choi2011map}. Some of the advantages of BIRL, compared to many other IRL algorithms, are (1) it finds a distribution over likely reward functions, (2) $D$ can contain partial demonstrations or even non-contiguous state action pairs, and (3) it works with sub-optimal demonstrations.

The choice of the prior allows domain knowledge to be inserted into the IRL algorithm. Ramachandran et al. \shortcite{ramachandran2007bayesian} give several possibilities such as a uniform, Gaussian, or Beta prior. For the remainder of this paper we assume the prior is uniform. Evaluating the effects of alternative priors is left to future work.



\section{Problem Definition} \label{sec:ProblemFormulation}
We assume that we are given an MDP$\setminus$R and samples $D = \{(s_1,a_1),\ldots,(s_m,a_m) | (s_i, a_i) \sim \pi_{\rm demo} \}$ of state-action pairs from a demonstrator's policy $\pi_{\rm demo}$. We make the common assumption \cite{abbeel2004apprenticeship,ramachandran2007bayesian} that the demonstrator attempts to maximize total return under the reward $R^*$ by executing a possibly sub-optimal, stationary policy $\pi_{\rm demo}$. Given any evaluation policy $\pi_{\rm eval}$, we are interested in the following general problem:

\subsubsection{High-confidence policy evaluation for LfD:} 
Given an MDP$\setminus$R, an evaluation policy $\pi_{\rm eval}$, and a set of demonstrations, $D$, find a high-confidence upper bound on the policy loss incurred by using $\pi_{\rm eval}$ in place of $\pi^*$, where $\pi^*$ is the optimal policy for the demonstrator's reward function, $R^*$.
\\

We define policy loss using the \textit{Expected Value Difference} (EVD) of $\pi_{\rm eval}$ under the true reward $R^*$, defined as  
\begin{equation}
\text{EVD}(\pi_{\rm eval}, R^*) = V^{*}_{R^*} - V^{\pi_{\rm eval}}_{R^*}.
\end{equation}
We use EVD because it is a natural way to measure the performance difference between two policies and it is a common metric for evaluating IRL algorithms \cite{ramachandran2007bayesian,levine2011nonlinear,choi2011map,wulfmeier2015maximum}. 
Note that the evaluation policy can be any policy, including a hand-tuned policy or a policy learned through reinforcement learning on a different task with a known reward function; however, the most natural form of the evaluation policy is a policy learned from the demonstrations, $D$.

We seek to bound the difference in expected return between the evaluation policy $\pi_{\rm eval}$ and $\pi^*$, the policy that is optimal with respect to the demonstrator's reward $R^*$.
However, because an optimal policy is invariant to any non-negative scaling of the reward function, bounding EVD is ill-posed, as we can multiply the feature weights $w$ by any $c>0$ to scale EVD to be anywhere in the range $[0,\infty)$. To avoid this scaling issue we make the common assumption that $\|w\|_1=1$ \cite{syed2008game,pirotta2016inverse}. Note, that this assumption only eliminates the trivial all-zero reward function as a potential solution---all other reward functions can be appropriately normalized. While setting $\|w\|_1 = 1$ eliminates the invariance to scaling factors and bounds the magnitude of the EVD, there can still be infinitely many rewards that induce any optimal policy, resulting in infinitely many possible values of $\text{EVD}(\pi_{\rm eval}, R^*)$. 
Thus, to obtain an upper bound on $\text{EVD}(\pi_{\rm eval}, R^*)$ we need to address this uncertainty. 

As we show in the following section, one way to address this uncertainty over the demonstrator's true reward is to compute an absolute worst-case policy loss bound using feature counts. However, as we show in the evaluation section, this type of worst-case bound is sensitive to adversarial reward functions that are highly unlikely given the demonstrations, often resulting in loose bounds.
Thus, rather than focusing on absolute worst-case, we focus on computing a probabilistic upper bound on the $\alpha$-worst-case value of $\text{EVD}(\pi_{\rm eval}, R)$, where $R \sim P(R|D)$. 

The $\alpha$-worst-case value of a random variable is often referred to in finance as the $\alpha$-Value at Risk \cite{jorion1997value}. We use the notation of Tamar et al. \cite{tamar2015optimizing} and formally define the \textit{$\alpha$-Value-at-Risk} of a random variable $Z$ as 
\begin{equation} \label{eqn:VaR}
\nu_{\alpha}(Z) = F^{-1}_Z(\alpha) = \inf \{ z : F_Z(z) \geq \alpha \}
\end{equation}
where $\alpha \in (0,1)$ is the quantile level and $F_Z(z) = Pr(Z \leq z)$ is the cumulative distribution function of $Z$. 

The specific problem that we address is the following:
\subsubsection{Risk-aware policy evaluation for LfD:}  Given an MDP$\setminus$R, any evaluation policy $\pi_{\rm eval}$, and a set of demonstrations, $D$, find a $(1-\delta)$ confidence upper bound on $\nu_{\alpha}(\text{EVD}(\pi_{\rm eval}, R))$, where $R\sim P(R|D)$.
\\

Note that $\alpha$ defines the sensitivity to risk, while $(1 - \delta)$ represents our confidence in our estimate of the $\alpha$-VaR. Thus, while $(1-\delta)$ is typically always high (e.g., 0.95), $\alpha$ can take on a range of values depending on the possibility of catastrophic failure in the domain and the risk-aversion of the end-user. In practice, $\alpha\geq0.9$ is commonly used for VaR applications \cite{jorion1997value}.

\section{Worst-Case Bound} \label{sec:MethodsForBoundingReturnDifference}
Before we give the full details of our approach, we first derive a simple worst-case bound based on feature counts that we use as a baseline. As a reminder, we use the notation $\mu(\pi) = \mathbb{E}_{s_0 \sim S_0}[\sum_{t=0}^{\infty}\gamma^t \phi(s_t) | \pi]$
to represent the expected feature counts of policy $\pi$.

%



Given any evaluation policy $\pi_{\rm eval}$, Abbeel and Ng \shortcite{abbeel2004apprenticeship} showed that if we assume $\phi(s):S\rightarrow[0,1]^k$, $\|w\|_1 \leq 1$, and know the demonstrator's expected feature counts $\mu^* = \mu(\pi_{\rm demo})$, then $\|\mu^* - \mu(\pi_{\rm eval})\|_2 \leq \epsilon$ implies that $$V^{\pi_{\rm demo}}_R - V^{\pi_{\rm eval}}_R = w^T (\mu^* - \mu(\pi_{\rm eval})) \leq \epsilon$$ for any reward function $R(s) = w^T \phi(s)$. 
If $\pi_{\rm demo}$ is optimal with respect to the demonstrator's reward function, $R^*$, then $$w^T (\mu^* - \mu(\pi_{\rm eval})) = \text{EVD}(\pi_{\rm eval}, R^*) \leq \epsilon$$ and  $\|\mu^* - \mu(\pi_{\rm eval})\|_2$ gives an upper bound on $\text{EVD}(\pi_{\rm eval}, R^*)$. 

We now derive an even tighter bound. First, note that the worst-case policy loss is the objective value of the following maximization problem
\begin{eqnarray}  
&\max_w& w^T(\mu^* - \mu(\pi_{\rm eval}))\\
&\text{subject to}& \|w\|_1 = 1.
\end{eqnarray}

The solution is to put all of our budget for $w$ on the feature with maximal feature count difference, giving the solution $\|\mu^* - \mu(\pi_{\rm eval})\|_\infty$. Because the two-norm is always lower bounded by the infinity-norm, this bound will be tighter than the bound proposed by Abbeel and Ng \shortcite{abbeel2004apprenticeship}. 

Note that in practice we do not know $\mu^*$, but we can use demonstrated trajectories to estimate of the demonstrator's expected feature counts as
\begin{equation}
\hat{\mu}^* = \frac{1}{|D|} \sum_{i=1}^{|D|} \sum_{t=0}^\infty \gamma^t \phi(s_t^{(i)}),
\end{equation}
where $i$ indexes over the trajectories and $t$ over the state sequence contained in each demonstrated trajectory. We define the empirical \emph{worst-case feature count bound} as 
\begin{equation} \label{eqn:wfcb}
\text{WFCB}(\pi_{\rm eval}, D) = \|\hat{\mu}^* - \mu(\pi_{\rm eval})\|_\infty.
\end{equation}

Note that for this bound to be a guaranteed upper bound on $\text{EVD}(\pi_{\rm eval}, R^*)$, $\pi_{\rm demo}$ must be optimal and the empirical estimate of the expert's feature counts, $\hat{\mu}^*$, may require a large number of demonstrations to converge to $\mu^*$ \cite{abbeel2004apprenticeship,syed2008game}.   
Other limitations of this bound are that it does not work with partial demonstrations and that it is based on an adversarial reward function
that may be extremely unlikely given the demonstrations.
%


\section{EVD Value-at-Risk Bound} \label{sec:OurMethod}
The worst-case feature count bound described in the previous section only requires sampled trajectories from the expert, but completely ignores the structure of the problem and the actions taken by the demonstrator---giving a worst-case bound that will likely be overly pessimistic. 
Our goal is to obtain a high-confidence probabilistic worst-case bound that focuses on likely reward functions given the demonstrations.

We seek a probabilistic confidence bound on the $\alpha$-Value at Risk of the EVD$(\pi_{\rm eval}, R^*)$ for any given evaluation policy $\pi_{\rm eval}$. 
We note that using the EVD rather than a standard feature count bound, as discussed in the previous section, is desirable for two main reasons. The first reason is that it works well with partial, noisy demonstrations. This is because EVD compares the evaluation policy against the optimal policy for reward $R$, not the actual states visited by the potentially sub-optimal demonstrator. Second, the EVD explicitly takes into account the initial state distribution. Thus, EVD measures the generalizability error of an evaluation policy by evaluating the expected return over all states with support under $S_0$, even if demonstrations have only been sampled from a small number of possible initial states.

To bound the $\alpha$-quantile worst-case $\text{EVD}(\pi_{\rm eval}, R^*)$ we use samples from the posterior $P(R|D)$. Thus, we seek to calculate $\nu_{\alpha}(Z)$ where $Z = EVD(\pi_{\rm eval}, R)$ for $R \sim P(R|D)$. 
As motivated in the Problem Definition, we assume $\|w\|_1 = 1$. Thus, to find $P(R|D)$ we use a modified version of the  BIRL Policy Walk Algorithm \cite{ramachandran2007bayesian} that ensures that our proposal samples of $w$ during MCMC stay on the L1-norm unit ball. Details are given in the Appendix. 
Using MCMC, we generate a sequence of sampled rewards $\mathcal{R}= \{R : R \sim P(R|D) \}$ from the posterior distribution over reward functions given the demonstrations.
For each sample $R_i \in {R}$ we then calculate \begin{equation}
Z_i = \text{EVD}(\pi_{\rm eval}, R_i) = V^{*}_{R_i} - V^{\pi_{\rm eval}}_{R_i}
\end{equation}
giving us samples from the posterior distribution over expected value differences. 

To obtain a point estimate of $\alpha$-VaR we can sort the resulting samples of $Z$ in ascending order to obtain the order statistics $Y$, and then take the $\alpha$-quantile. However, this does not take into account the number of samples or our confidence in this point estimate. Instead of using a point estimate, we compute a single-sided $(1-\delta)$ confidence bound on the $\alpha$-VaR. Given a sample $Z_i$, we have that $P(Z_i  < \nu_{\alpha}(Z)) = \alpha$. Thus, for any order statistic $Y_j$, we can use the normal approximation of the binomial distribution to obtain
\begin{eqnarray} \label{eqn:binomial}
\displaystyle P(\nu_{\alpha}(Z) \leq Y_j) &=& \sum_{i=1}^{j} \binom{N}{i} \alpha^i (1-\alpha)^{N-i} \\ &\approx& F_\mathcal{N}\bigg(j + \frac{1}{2} \mid N \alpha, N \alpha (1- \alpha)\bigg) .  \nonumber
\end{eqnarray}
where $F_\mathcal{Z}$ is the CDF of the normal distribution with $\mu = N \alpha$ and $\sigma^2 = N \alpha (1- \alpha)$ and 1/2 is added to the index $j$ as a continuity correction \cite{hollander1999nonparametric}. To obtain the index $k$ of the order statistic such that $P(\nu_{\alpha}(Z) \leq Y_k) \geq (1-\delta)$ we invert Equation~\ref{eqn:binomial} using the inverse of the standard normal CDF, $F^{-1}_{ \mathcal{N}}$, to get $k =  \lceil N \alpha +  F^{-1}_{ \mathcal{N}}(1-\delta) \sqrt{N \alpha (1 - \alpha)} - \frac{1}{2}  \rceil$. Our full approach is summarized in Algorithm~\ref{alg:BIRL_bound}. The algorithm has three hyperparameters: $c$ defines the confidence in the optimality of the demonstrations, $\alpha$ defines the risk-sensitivity, and $(1 - \delta)$ represents the desired confidence level on the estimate of the $\alpha$-VaR.

\begin{algorithm}[t]
\caption{$(1-\delta)$ Confidence Bound on the $\alpha$-Value-at-Risk}\label{alg:BIRL_bound}
\begin{algorithmic}[1]
\State \textbf{input:} MDP$\setminus$R, $\pi_{\rm eval}$, $D$, $c$, $\alpha$, $\delta$ 
\State $\mathcal{R} \gets$ \textbf{BIRL}(MDP$\setminus$R, $D$, $c$) \Comment sample from posterior using L1-unit norm walk 
\For{$R_i \in \mathcal{R}$} 
\State $Z_i  = V^*_{R_i} - V^{\pi_{\rm eval}}_{R_i} $ \Comment compute policy loss
\EndFor
\State $Y = $ sort($Z$) \Comment sort into ascending order statistics
\State $k =  \lceil N \alpha +  F^{-1}_{ \mathcal{N}}(1-\delta) \sqrt{N \alpha (1 - \alpha)} - \frac{1}{2}  \rceil$ \Comment index of $(1-\delta)$ confidence bound on $\alpha$-VaR
\State \Return $Y_k$
\end{algorithmic}
\end{algorithm}

%

The advantages of our approach are as follows: (1) our proposed bound takes full advantage of all of the information contained in the transition dynamics and demonstrations to focus on reward functions that are likely given the demonstrations, (2) it does not require optimal demonstrations, (3) it inherits from BIRL the ability to work with partial demonstrations, even disjoint state-action pairs, and (4) it allows for domain knowledge in the form of a prior.

\section{Empirical results} \label{sec:Experiments}

\begin{figure}
\centering
\subfigure[Grid world navigation]{\includegraphics[scale=0.12]{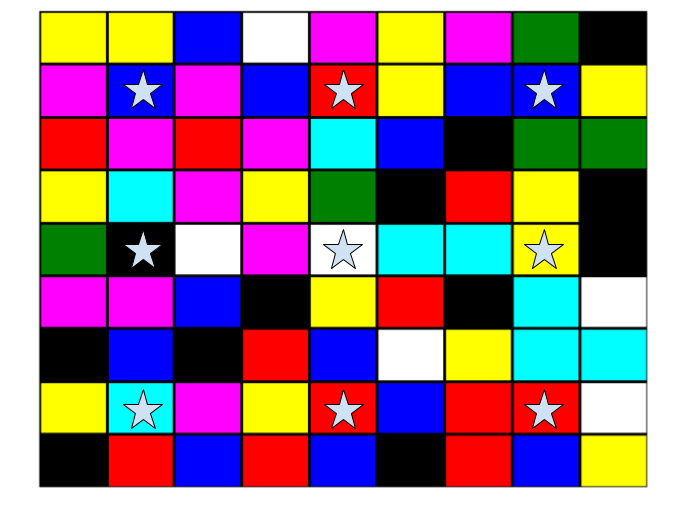}
\label{subfig:grid}}
\subfigure[Driving simulation]{
\includegraphics[scale=0.155]{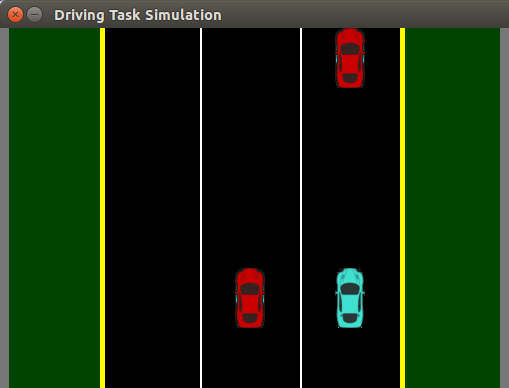}
\label{subfig:drivingWorld}}
\caption{(a) Example of random grid world navigation task with colors representing random features and initial states denoted by stars. (b) Snapshot of driving simulation. Agent must learn to safely drive blue car through traffic.}
\label{fig:tasks}
\end{figure}

For our proposed confidence bound to be useful, it needs to meet several criteria: (1) the upper bound should be accurate with high-confidence (rarely underestimating the true expected value difference), (2) the bound should be tighter than the worst-case bound derived above, and (3) the previous two criteria should be true even when given a small number of demonstrations. We use both a standard grid world navigation task \cite{abbeel2004apprenticeship,ramachandran2007bayesian,choi2011map} and a simulated driving task \cite{abbeel2004apprenticeship,syed2008game,cohn2011comparing} to validate that our proposed bound satisfies these criteria. Examples of these tasks are shown in Figure~\ref{fig:tasks}. We compare our high-confidence $\alpha$-VaR bound with the worst-case feature count bound (WFCB) defined in Equation \ref{eqn:wfcb}. All results for $\alpha$-VaR bounds are reported as 95\% confidence bounds ($\delta = 0.05$). 




\subsection{Grid world navigation task}
We first empirically evaluate our approach on a suite of 9x9 grid world navigation tasks where the cost of traveling on different terrains is unknown and must be inferred from demonstrations. The available actions are up, down, left and right. Transitions are noisy with an 70\% chance of moving in the desired direction and 30\% chance of going in one of the directions perpendicular to the chosen direction. There are 8 binary features with one feature active per grid cell. To show that our results are not an artifact of a specific reward function, we evaluate our method over many random grid worlds, each with a randomly chosen ground truth reward. We use $\gamma =0.9$ and an initial state distribution $S_0$ that is uniform over 9 different states spread across the grid as shown in Figure~\ref{subfig:grid}. When generating $M$ demonstrations we select the initial states in a round-robin fashion from the support of $S_0$. However, when measuring accuracy and bound errors, we compare with the true expected value difference over the full initial state distribution.

\subsubsection{Infinite horizon grid navigation} \label{sec:noTerminals}
Our first task is an infinite horizon grid world navigation task with no terminal states. To evaluate different bounding methods we generated 200 random 9x9 worlds with random features each grid cell. For each world we generated a random feature weight vector $w$ from the L1-unit norm ball. To generate demonstrations we solve the MDP using the random ground truth reward to find the optimal policy and use this policy to generate trajectories of length 100.
We set the evaluation policy to be the optimal policy under the MAP reward function found using BIRL. Because the demonstrations in this experiment are perfect, we set the BIRL confidence parameter to a large value ($c = 100$). 

\begin{figure}[t]
\centering
\subfigure[Accuracy]{\includegraphics[scale=0.31]
{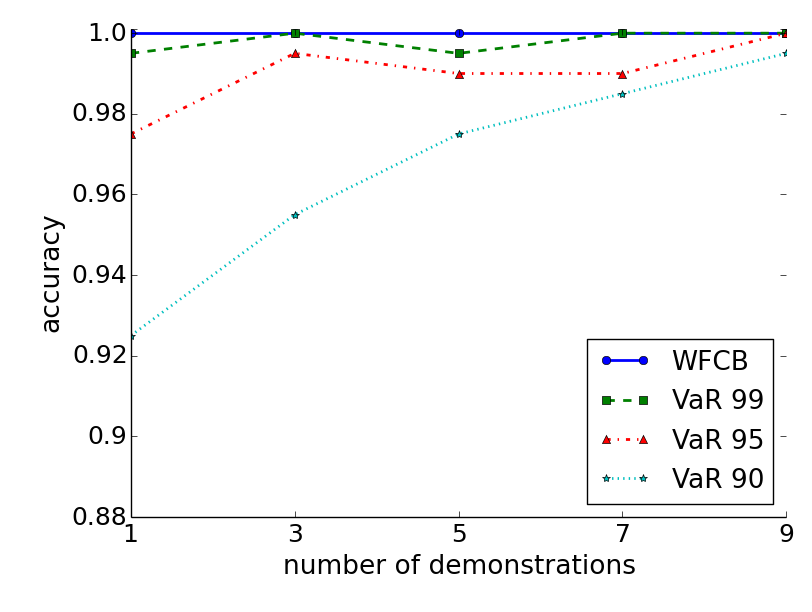}
\label{subfig:NoTermNoDupAccuracy}}
\subfigure[Average Bound Error]{
\includegraphics[scale=0.31]
{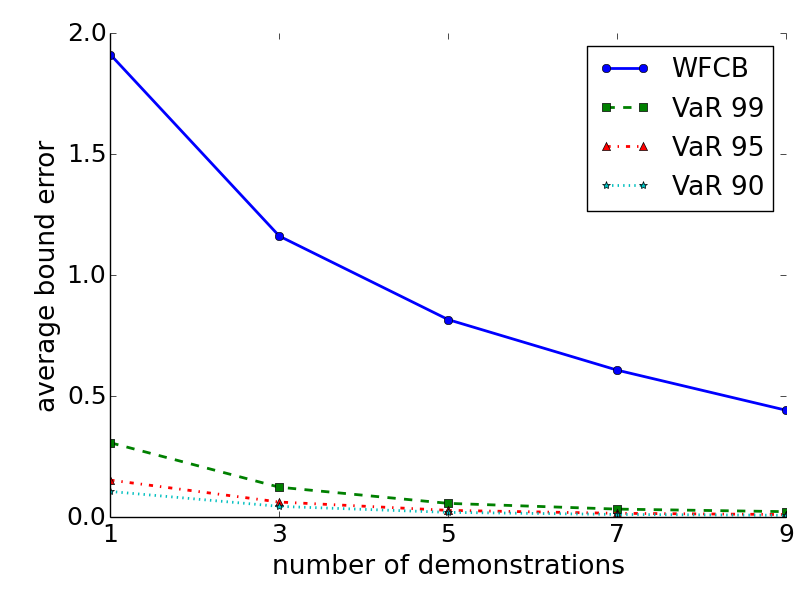}
\label{subfig:NoTermNoDupBoundError}}
\caption{Results for infinite horizon grid navigation task. Accuracy and average error for bounds based on feature counts (WFCB) compared with 99, 95, and 90 percentiles for the VaR bound.  Accuracy and averages are computed  over 200 replicates}
\label{fig:gridNoTerminalNoDuplicates}
\end{figure}

Figure~\ref{subfig:NoTermNoDupAccuracy} shows the accuracy of each bound where WFCB is the worst-case feature count bound, and VaR X is the X/100 quantile Value at Risk bound. The accuracy is the proportion of trials where the upper bound is greater than the ground truth expected value difference over the 200 random grid worlds. As expected, the WFCB always gives an upper bound on the true performance difference between the optimal policy and the evaluation policy. The bounds on $\alpha$-VaR are also highly accurate.

Because always predicting a high upper bound will result in high accuracy, we also measured the tightness of the the upper bounds. Figure~\ref{subfig:NoTermNoDupBoundError} shows the average bound error over the 200 random navigation tasks. We define the bound error for an upper bound $b$ as \begin{equation}\label{eqn:boundError}
\text{error}(b) = b - \text{EVD}(\pi_{\rm eval}, R^*)
\end{equation}
where $R^*$ is the generated ground truth reward. We see that the bounds on the $\alpha$-VaR are much tighter than the worst-case feature count bound, converging after only a small number of demonstrations.

\subsubsection{Noisy demonstrations}
\begin{figure}[t]
\centering
\subfigure[Accuracy]{\includegraphics[scale=0.28]
{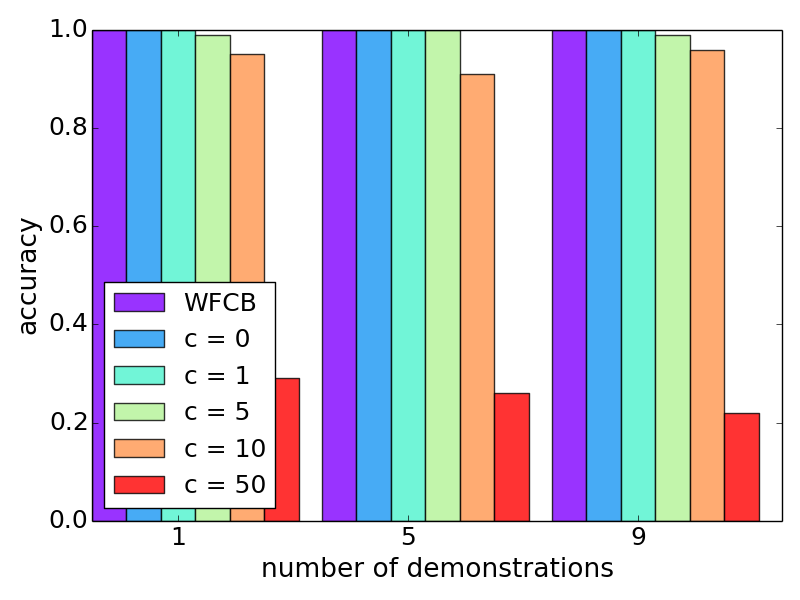}
\label{subfig:NoTermNoisyTAccuracy}}
\subfigure[Average Bound Error]{
\includegraphics[scale=0.28]
{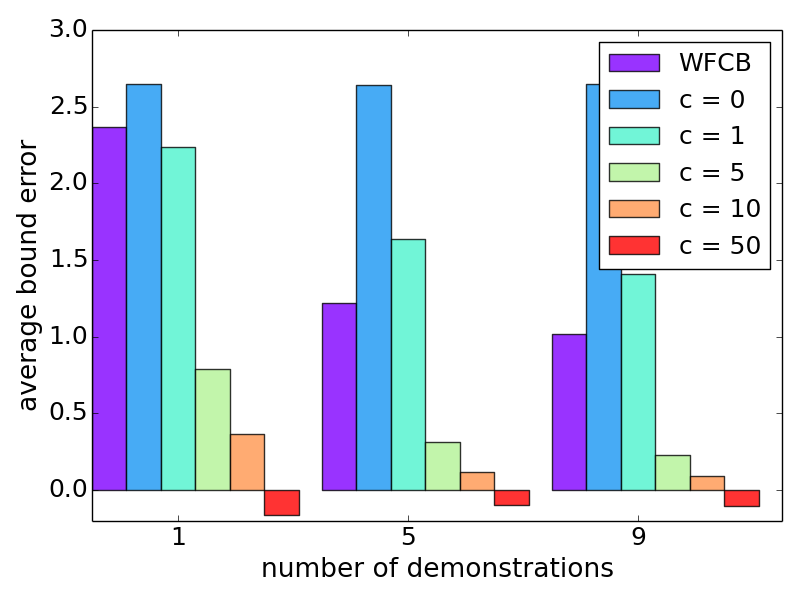}
\label{subfig:NoTermNoisyTBoundError}}
\caption{Sensitivity to the confidence $c$ for noisy demonstrations in the grid navigation task. The demonstrator has a 20\% chance of taking a random action in each state. Accuracy and average error for bounds based on feature counts (WFCB) compared with 0.95-VaR bound. Accuracy and averages are computed  over 200 replicates.}
\label{fig:noisyDemos}
\end{figure}

As mentioned previously, BIRL uses a confidence parameter, $c$, that represents the optimality of the
demonstrations. When $c = 0$, the demonstrations are assumed to come from a completely random
policy, and $c = \infty$  means that the demonstrations come from a perfectly optimal policy. Prior work used values of $c$ between 25 and 500 when demonstrations are generated from an
expert policy \cite{lopes2009active,cohn2011comparing,michini2012improving}. To investigate the effect of $c$ on our bound we generated noisy demonstrations where at step there is an 80\% chance of taking an optimal action and a 20\% chance of taking a random action. The resulting accuracy and bound error for several choices of $c$ are shown in
Figure~\ref{fig:noisyDemos}.

Adjusting $c$ for noisy demonstrations has a clear effect on the accuracy and bound error. The bound error (Equation~\ref{eqn:boundError}) decreases as $c$ increases, meaning the bounds become tighter; however, when $c=50$ the VaR bounds often underestimate the true expected value difference between the
expert's policy and the evaluation policy, resulting in error$(b)<0$ and lower accuracy. We see that values of $c$ in the range $(1,10]$ result in highly accuracy bounds that are tighter
than the worst-case feature count bound. However, for $c=50$, we see that BIRL overfits to the noise in the demonstrations by assuming that the demonstrations are optimal. Tuning the confidence parameter, $c$, for a particular demonstrator and task is left for future work.

\subsubsection{Evaluation policy}
In the previous examples we have used the MAP reward obtained from BIRL to create the evaluation policy; however, unlike previous theoretical confidence bounds, our method is applicable to any evaluation policy. 
We investigated the sensitivity of our bound over a range of different evaluation policies and found that the VaR bounds consistently outperforms the baseline WFCB, providing bounds that are often four times tighter while maintaining high accuracy (see the Appendix for details). 

To demonstrate the ability of our method to work with evaluation policies derived from other IRL algorithms, and to compare against existing high-confidence bounds for IRL, we used the  Projection algorithm proposed by Abbeel and Ng \cite{abbeel2004apprenticeship} as an evaluation policy. Abbeel and Ng provide high-confidence bounds on the number of demonstrations needed for their algorithm to guarantee performance within $\epsilon$ of the demonstrator. A tighter sample bound for feature count-based methods was later derived by Syed and Schapire \shortcite{syed2008game} that also holds for the Projection algorithm. We inverted the bound of Syed and Schapire to obtain a $(1-\delta)$ confidence bound on the expected value difference given a fixed number of demonstrations (see the Appendix for details).

We then repeated the infinite horizon grid navigation experiment described above, using the policy found by the Projection algorithm as our evaluation policy. We compare the average bound error for our proposed VaR bounds with the Syed and Schapire error bound for the Projection algorithm in Table~\ref{tab:AbbeelBound}. Our empirical VaR bounds are two to three orders of magnitude tighter than the Hoeffding style bound which theoretically requires 23,146 demonstrations to guarantee the true EVD is within the 0.95-VaR bound found by our method using only a single demonstration.

\begin{table*}[t]

  \centering
  \begin{tabular}{cccccccc}
    \toprule
    & \multicolumn{5}{c}{Number of demonstrations}                  & Average Accuracy \\
    \cmidrule{2-6}
     & 1 & 5 & 9 & $\cdots$ & 23,146 &\\
    \midrule
    0.95-VaR EVD Bound& \textbf{0.9372} & \textbf{0.2532} & \textbf{0.1328} &  & - & 0.98\\
   0.99-VaR EVD Bound & 1.1428 &  0.2937 & 0.1535 & & - &1.0 \\
     EVD Bound \cite{syed2008game} & 142.59 & 63.77 & 47.53 & & 0.9372 & 1.0\\ 
    \bottomrule
  \end{tabular}
   \caption{Comparison of 95\% confidence $\alpha$-VaR bounds with a 95\% confidence Hoeffding-style bound \cite{syed2008game}. Both bounds use the Projection algorithm \cite{abbeel2004apprenticeship} to obtain the evaluation policy. Results are averaged over 200 random navigation tasks.}
   \label{tab:AbbeelBound}

\end{table*}

\subsection{Driving task}
We now provide an example that more closely matches a real-world learning from demonstration task. Rather than evaluate our method on an ad hoc ``true" reward function, we examine how the VaR bound can be used to rank and select an appropriate policy from a set of existing policies.
For this task we designed a driving simulator based on previous benchmarks \cite{abbeel2004apprenticeship,cohn2011comparing}. Figure~\ref{subfig:drivingWorld} shows a snapshot of the simulator. The agent (blue) is in charge of driving safely down a highway and has three actions: switch lanes left, switch lanes right, or stay in current lane. The agent is traveling faster than traffic and must change lanes to avoid other cars which randomly appear at the top of the screen. There are three highway lanes where the car is supposed to drive, but it can also drive off-road on the right or left of the highway.

The state space is made up of 12 binary features: 5 features for each of the possible lanes, including the off-road lanes, 3 features telling the agent whether it is currently in collision, tailgating, or trailing another car, and 2 features for each adjacent lane, indicating whether the car will be in collision or tailgating if the car changes lanes. The reward is assumed to be a linear combination of features, $R(s) = w^T \phi(s)$, where $\phi(s)$ is a 6-dimensional binary feature vector that indicates the agent's current lane and whether it is in collision with another car. The discount factor, $\gamma$, was set to 0.9.

The goal of this experiment is to evaluate the ability of our probabilistic performance bound to correctly rank different policies, given a single demonstration of safe driving. We constructed three different evaluation policies: (1) \textbf{right-safe}: a policy that avoids hitting cars and driving off-road, but prefers driving on the right lane of the highway, (2) \textbf{on-road}: a policy that avoids driving off-road, but pays no attention to other cars, and changes lanes randomly (3) \textbf{nasty}: a policy that avoids going off-road, but actively tries to hit cars.
We then generated a single demonstration of collision-free driving, consisting of 100 consecutive state-action pairs. The demonstration changed lanes randomly while avoiding collisions and avoiding driving off-road. The evaluation policies and demonstration were created using Q-learning and hand-crafted reward functions that resulted in the desired behaviors. 

Because the driving task is model-free we used Q-learning to calculate the Q-values used in the likelihood calculations of BIRL. We then calculated a 95\% confidence bound on the 0.95-VaR for each evaluation policy. We also computed the worst-case feature count bounds for comparison. The results are shown in Table~\ref{tab:driving}.

The VaR bound uses the demonstration to focus on reward functions that are likely given the demonstrated state-action pairs. This results in correctly ranking the evaluation policies. The worst-case feature count bound ignores likelihood and assumes a worst-case reward function that penalizes the largest discrepancy between the empirical feature counts of the demonstration and the expected feature counts of the evaluation policies. Because the collision feature is less frequently active than the lane features, both \textbf{on-road} and \textbf{nasty} appear safer than \textbf{right-safe} because their average state-occupancies more closely align with the state-occupancies of the demonstration.






\begin{table}
\centering
\begin{tabular}{cccll}
    \toprule
& & \multicolumn{3}{c}{Ranking (EVD upper bound)} \\
 \cmidrule{3-5}
$\pi_{\rm eval}$ & Collisions & True & WFCB & 0.95-VaR \\ 
\midrule
right-safe  & 0 & \textbf{1} & 3 (5.51) & \textbf{1} (0.85) \\  
on-road & 13.65 & \textbf{2} & 1 (1.93) & \textbf{2} (1.09) \\ 
nasty  & 42.75 & \textbf{3} & 2 (4.11) & \textbf{3} (2.44)\\ 
\bottomrule
\end{tabular}
\caption{Policy rankings based on upper bounds on policy loss for three different evaluation policies in the driving domain when given a single demonstration of safe driving. Results are averaged over 20 replicates.}
\label{tab:driving}
\end{table}


\subsection{High-confidence policy improvement}
The previous section showed how we can use risk-sensitive policy evaluation to choose between multiple evaluation policies. We now take this a step further and give an example that uses risk-sensitive policy evaluation to iteratively reduce the VaR of a policy learned from demonstrations.

\begin{figure}
\centering
\includegraphics[scale=0.12]{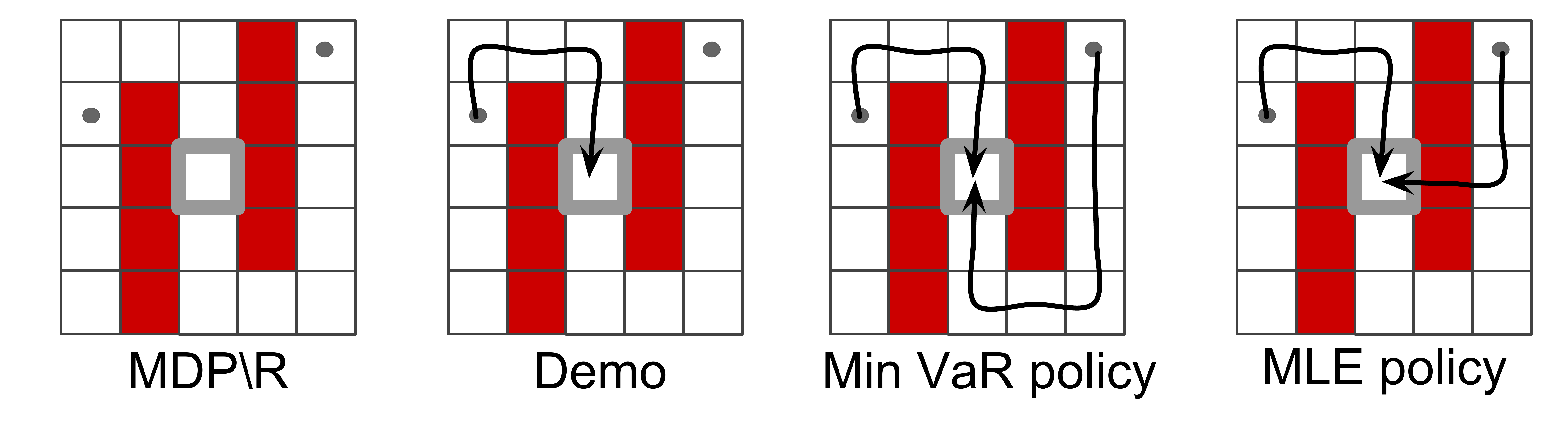}
\caption{Given one demonstration, optimizing the VaR bound results in a risk-aware policy that hedges against the red cells being much worse than the white. The maximum likelihood reward assumes that red is only marginally worse than white.}
\label{fig:safeImprovement}
\end{figure}

To highlight the potential of safe policy improvement, we consider the simple navigation task shown in Figure~\ref{fig:safeImprovement}. The task has a single terminal in the center and two reward features (white and red). The agent is given a single demonstration from one starting state and must generalize this demonstration to a second starting state (both marked with circles). Note that the demonstration shows that the red feature is less desirable than the white feature, but the true magnitudes of the feature weights are left uncertain.

We implemented a simple risk-sensitive policy improvement hill climbing algorithm. We initialized the hill climbing algorithm with the maximum likelihood policy found using BIRL with a uniform prior. For each step of the hill-climbing algorithm, we examined the impact on the 0.99-VaR of changing the action taken by the policy in a single state, and chose the change that resulted in the largest decrease in 0.99-VaR over all single state changes. We continued this process until no reductions in the 0.99-VaR could be found. The resulting risk-aware policy seeks to minimize the 0.99-VaR by avoiding the red feature, whereas the maximum likelihood reward leads to a less conservative policy, resulting in a higher potential risk. The learned policies are shown in Figure~\ref{fig:safeImprovement}. In the future, more complex policy adaptation schemes such as finite difference methods or black-box optimization techniques (e.g. CMA-ES \cite{hansen2006cma}) could also be used to approximate the gradient of the $\alpha$-VaR with respect to a parameterized policy $\pi$.

\section{Related work} \label{sec:RelatedWork}
Many different methods exist for learning from demonstration through inverse reinforcement learning \cite{Argall2009,gao2012survey}. However, few of them give guarantees on performance. Abbeel and Ng \shortcite{abbeel2004apprenticeship} and Syed and Schapire \shortcite{syed2008game} give probabilistic Hoeffding-style bounds on how many demonstrations their algorithms require to guarantee a policy with expected return within epsilon of the expected return of the demonstrator's policy. However, as shown in Table~\ref{tab:AbbeelBound}, these theoretical bounds are too loose to be useful in practice and are customized for their specific IRL algorithms. To our knowledge, we provide the first sample-efficient, high-confidence bound on the policy loss of any evaluation policy with respect to the optimal policy under the demonstrator's true reward function.


Safety has been extensively studied within the reinforcement learning community (see Garcia et al. \shortcite{garcia2015comprehensive} for a survey). These approaches typically either focus on safe exploration or on optimizing an objective other than expected return. Recently, alternative objectives based on financial measures of risk such as VaR and Conditional VaR have been shown to provide tractable and useful risk-sensitive measures of performance for MDPs \cite{tamar2015optimizing,chow2015risk}. Santara et al. \shortcite{santara2017rail} propose an algorithm to minimize conditional VaR for generative adversarial imitation learning, but do not provide bounds on the safety of the learned policy. Our work complements prior research on safety in reinforcement learning and imitation learning by showing how risk-sensitive metrics can be applied to IRL to obtain high-confidence performance bounds.

Additional work on safety in MDPs has focused on obtaining high-confidence bounds on the performance of a policy before that policy is deployed \cite{thomas2015high,hanna2017bootstrapping}, as well as methods for high-confidence policy improvement \cite{thomas2015improvement}. 
Our work draws inspiration from these previous approaches; however, we provide bounds on policy performance that are applicable when learning from demonstrations, i.e., when the rewards are not observed.


%
%
%

\section{Discussion and Future Work}
Due to space and time constraints we did not explore the full range of possible instantiations of a risk-sensitive performance bound for learning from demonstration through IRL. In this section we discuss design choices, limitations of our approach, and avenues for future research.

We decided to measure policy loss using EVD as it is a commonly used IRL metric; however, this is not the only measure of performance that can be used in our approach. Because our method estimates the posterior distribution over reward functions, any risk measure or loss that is a function of a reward function and a policy can be inserted into our framework in place of EVD.

We used VaR because it well known and widely used, easy to implement using Monte-Carlo samples, and is a probabilistic analogue to the WFCB. However, our proposed methodology can be extended to use other risk measure that can be computed from samples of a distribution. Alternative risk measures such as Conditional Value-at-Risk \cite{rockafellar2000optimization}, entropic risk measure \cite{follmer2011entropic}, or semideviations \cite{ogryczak1999stochastic} could replace VaR in our framework. Recently, methods have been proposed that explicitly optimize the Conditional VaR of a policy \cite{tamar2015optimizing,santara2017rail}. Future work should examine whether these approaches can be combined with our work on risk-aware policy improvement for IRL.

Because our bound is based on Bayesian IRL, our method is designed to work with partial demonstrations and allows insertion of domain knowledge as a prior over reward functions. \citeauthor{choi2011map}~\shortcite{choi2011map} have shown that many standard IRL algorithms can be transformed into an equivalent Bayesian IRL algorithm by selecting the appropriate likelihood and prior. Thus, our proposed performance bound can be easily extended to use alternative likelihoods and priors that match different assumptions and preferences found in the IRL literature.

One of the main drawbacks of our proposed framework is that it requires running MCMC, which repeatedly samples rewards and then solves for $Q^*_R(s,a)$ and $V^*_R$ in order to calculate the BIRL likelihood and to compute samples of $\text{EVD}(\pi_{\rm eval}, R^*)$. Future work should investigate whether IRL methods based on policy gradients  \cite{pirotta2016inverse,ho2016model} or other IRL algorithms that do not require repeatedly solving an MDP \cite{boularias2011relative,kalakrishnan2013learning,finn2016guided} can be used to sample from the posterior distribution over reward functions.

Our method also relies on an appropriate range for the confidence parameter $c$ in the BIRL algorithm, which determines how much we trust the demonstrations. Recently, an Expectation Maximization approach has been used to learn this parameter from a large number of demonstrations of differing quality \cite{zheng2014robust}. Future work should investigate whether a similar approach can be used to learn an appropriate value for $c$ when there are possibly only a small number of demonstrations of similar quality.

\section{Conclusion} \label{sec:Conclusion}
In this work we have formalized and addressed the problem of risk-aware high-confidence policy evaluation with an unknown reward function. To our knowledge, we present the first general framework for obtaining practical high-confidence bounds on the performance difference between an evaluation policy and the optimal policy for a demonstrator's true unknown reward. We also give examples of how our high-confidence performance bound can be used to perform risk-aware policy selection and risk-aware policy improvement. Our proposed algorithms are evaluated on a standard grid navigation task and driving simulation.

Our results demonstrate that our proposed bound is a significant improvement over a baseline based on feature counts---providing accurate, tight bounds even for small numbers of demonstrations. Additionally, our empirical results show orders of magnitude improvement in sample efficiency over competing confidence bounds \cite{abbeel2004apprenticeship,syed2008game}. As a result, this is the first approach that allows agents that learn from demonstrations to express confidence in the performance of their learned policy, based on limited demonstration data. We believe the techniques proposed in this paper provide a starting point for developing autonomous agents that can safely and efficiently learn from human demonstrations in risk-sensitive, real-world environments.

\section{Acknowledgments}
This work has taken place in the Personal Autonomous Robotics
Lab (PeARL) at The University of Texas at Austin. PeARL research
is supported in part by the NSF (IIS-1638107, IIS-1617639, IIS-
1724157).

\fontsize{9.0pt}{10.0pt} 
\selectfont

\bibliographystyle{aaai}
\bibliography{nipsRefs}

\begin{thebibliography}{}

\bibitem[\protect\citeauthoryear{Abbeel and
  Ng}{2004}]{abbeel2004apprenticeship}
Abbeel, P., and Ng, A.~Y.
\newblock 2004.
\newblock Apprenticeship learning via inverse reinforcement learning.
\newblock In {\em Proceedings of the 21st international conference on Machine
  learning}.

\bibitem[\protect\citeauthoryear{Amodei \bgroup et al\mbox.\egroup
  }{2016}]{amodei2016concrete}
Amodei, D.; Olah, C.; Steinhardt, J.; Christiano, P.; Schulman, J.; and
  Man{\'e}, D.
\newblock 2016.
\newblock Concrete problems in ai safety.
\newblock {\em arXiv preprint arXiv:1606.06565}.

\bibitem[\protect\citeauthoryear{Argall \bgroup et al\mbox.\egroup
  }{2009}]{Argall2009}
Argall, B.~D.; Chernova, S.; Veloso, M.; and Browning, B.
\newblock 2009.
\newblock A survey of robot learning from demonstration.
\newblock {\em Robotics and autonomous systems} 57(5):469--483.

\bibitem[\protect\citeauthoryear{Babes \bgroup et al\mbox.\egroup
  }{2011}]{babes2011apprenticeship}
Babes, M.; Marivate, V.; Subramanian, K.; and Littman, M.~L.
\newblock 2011.
\newblock Apprenticeship learning about multiple intentions.
\newblock In {\em Proceedings of the 28th International Conference on Machine
  Learning}.

\bibitem[\protect\citeauthoryear{Baker, Saxe, and
  Tenenbaum}{2009}]{baker2009action}
Baker, C.~L.; Saxe, R.; and Tenenbaum, J.~B.
\newblock 2009.
\newblock Action understanding as inverse planning.
\newblock {\em Cognition} 113(3):329--349.

\bibitem[\protect\citeauthoryear{Barthe \bgroup et al\mbox.\egroup
  }{2005}]{barthe2005probabilistic}
Barthe, F.; Gu{\'e}don, O.; Mendelson, S.; Naor, A.; et~al.
\newblock 2005.
\newblock A probabilistic approach to the geometry of the $\ell^n_p$-ball.
\newblock {\em The Annals of Probability} 33(2):480--513.

\bibitem[\protect\citeauthoryear{Bishop}{2006}]{bishop2006pattern}
Bishop, C.~M.
\newblock 2006.
\newblock {\em Pattern recognition and machine learning}.
\newblock springer.

\bibitem[\protect\citeauthoryear{Boularias, Kober, and
  Peters}{2011}]{boularias2011relative}
Boularias, A.; Kober, J.; and Peters, J.
\newblock 2011.
\newblock Relative entropy inverse reinforcement learning.
\newblock In {\em Proceedings of the Fourteenth International Conference on
  Artificial Intelligence and Statistics}.

\bibitem[\protect\citeauthoryear{Brown and Niekum}{2017}]{brown2017full}
Brown, D.~S., and Niekum, S.
\newblock 2017.
\newblock Efficient probabilistic performance bounds for inverse reinforcement
  learning.
\newblock {\em arXiv preprint arXiv:1707.00724}  (Full paper).

\bibitem[\protect\citeauthoryear{Brown \bgroup et al\mbox.\egroup
  }{2016}]{brown2016exact}
Brown, D.~S.; Hudack, J.; Gemelli, N.; and Banerjee, B.
\newblock 2016.
\newblock Exact and heuristic algorithms for risk-aware stochastic physical
  search.
\newblock {\em Computational Intelligence}.

\bibitem[\protect\citeauthoryear{Choi and Kim}{2011}]{choi2011map}
Choi, J., and Kim, K.-E.
\newblock 2011.
\newblock Map inference for bayesian inverse reinforcement learning.
\newblock In {\em Advances in Neural Information Processing Systems}.

\bibitem[\protect\citeauthoryear{Chow \bgroup et al\mbox.\egroup
  }{2015}]{chow2015risk}
Chow, Y.; Tamar, A.; Mannor, S.; and Pavone, M.
\newblock 2015.
\newblock Risk-sensitive and robust decision-making: a cvar optimization
  approach.
\newblock In {\em Advances in Neural Information Processing Systems}.

\bibitem[\protect\citeauthoryear{Cohn, Durfee, and
  Singh}{2011}]{cohn2011comparing}
Cohn, R.; Durfee, E.; and Singh, S.
\newblock 2011.
\newblock Comparing action-query strategies in semi-autonomous agents.
\newblock In {\em The 10th International Conference on Autonomous Agents and
  Multiagent Systems}.

\bibitem[\protect\citeauthoryear{Finn, Levine, and
  Abbeel}{2016}]{finn2016guided}
Finn, C.; Levine, S.; and Abbeel, P.
\newblock 2016.
\newblock Guided cost learning: Deep inverse optimal control via policy
  optimization.
\newblock In {\em International Conference on Machine Learning}.

\bibitem[\protect\citeauthoryear{F{\"o}llmer and
  Knispel}{2011}]{follmer2011entropic}
F{\"o}llmer, H., and Knispel, T.
\newblock 2011.
\newblock Entropic risk measures: Coherence vs. convexity, model ambiguity and
  robust large deviations.
\newblock {\em Stochastics and Dynamics} 11(02n03):333--351.

\bibitem[\protect\citeauthoryear{Gao \bgroup et al\mbox.\egroup
  }{2012}]{gao2012survey}
Gao, Y.; Peters, J.; Tsourdos, A.; Zhifei, S.; and Meng~Joo, E.
\newblock 2012.
\newblock A survey of inverse reinforcement learning techniques.
\newblock {\em International Journal of Intelligent Computing and Cybernetics}
  5(3):293--311.

\bibitem[\protect\citeauthoryear{Garc{\i}a and
  Fern{\'a}ndez}{2015}]{garcia2015comprehensive}
Garc{\i}a, J., and Fern{\'a}ndez, F.
\newblock 2015.
\newblock A comprehensive survey on safe reinforcement learning.
\newblock {\em Journal of Machine Learning Research} 16(1):1437--1480.

\bibitem[\protect\citeauthoryear{Hanna, Stone, and
  Niekum}{2017}]{hanna2017bootstrapping}
Hanna, J.~P.; Stone, P.; and Niekum, S.
\newblock 2017.
\newblock Bootstrapping with models: Confidence intervals for off-policy
  evaluation.
\newblock In {\em Proceedings of the 16th Conference on Autonomous Agents and
  Multiagent Systems}.

\bibitem[\protect\citeauthoryear{Hansen}{2006}]{hansen2006cma}
Hansen, N.
\newblock 2006.
\newblock The cma evolution strategy: a comparing review.
\newblock {\em Towards a new evolutionary computation}  75--102.

\bibitem[\protect\citeauthoryear{Ho, Gupta, and Ermon}{2016}]{ho2016model}
Ho, J.; Gupta, J.; and Ermon, S.
\newblock 2016.
\newblock Model-free imitation learning with policy optimization.
\newblock In {\em International Conference on Machine Learning},  2760--2769.

\bibitem[\protect\citeauthoryear{Hollander and
  Wolfe}{1999}]{hollander1999nonparametric}
Hollander, M., and Wolfe, D.~A.
\newblock 1999.
\newblock {\em Nonparametric Statistical Methods: By Myles Hollander, Douglas
  A. Wolfe}.
\newblock J. Wiley.

\bibitem[\protect\citeauthoryear{Jorion}{1997}]{jorion1997value}
Jorion, P.
\newblock 1997.
\newblock {\em Value at risk}.
\newblock McGraw-Hill, New York.

\bibitem[\protect\citeauthoryear{Kalakrishnan \bgroup et al\mbox.\egroup
  }{2013}]{kalakrishnan2013learning}
Kalakrishnan, M.; Pastor, P.; Righetti, L.; and Schaal, S.
\newblock 2013.
\newblock Learning objective functions for manipulation.
\newblock In {\em IEEE International Conference on Robotics and Automation},
  1331--1336.

\bibitem[\protect\citeauthoryear{Karasev \bgroup et al\mbox.\egroup
  }{2016}]{karasev2016intent}
Karasev, V.; Ayvaci, A.; Heisele, B.; and Soatto, S.
\newblock 2016.
\newblock Intent-aware long-term prediction of pedestrian motion.
\newblock In {\em IEEE International Conference on Robotics and Automation},
  2543--2549.

\bibitem[\protect\citeauthoryear{Kim and Pineau}{2016}]{kim2016socially}
Kim, B., and Pineau, J.
\newblock 2016.
\newblock Socially adaptive path planning in human environments using inverse
  reinforcement learning.
\newblock {\em International Journal of Social Robotics} 8(1):51--66.

\bibitem[\protect\citeauthoryear{Levine, Popovic, and
  Koltun}{2011}]{levine2011nonlinear}
Levine, S.; Popovic, Z.; and Koltun, V.
\newblock 2011.
\newblock Nonlinear inverse reinforcement learning with gaussian processes.
\newblock In {\em Advances in Neural Information Processing Systems}.

\bibitem[\protect\citeauthoryear{Lopes, Melo, and
  Montesano}{2007}]{lopes2007affordance}
Lopes, M.; Melo, F.~S.; and Montesano, L.
\newblock 2007.
\newblock Affordance-based imitation learning in robots.
\newblock In {\em IEEE/RSJ International Conference on Intelligent Robots and
  Systems},  1015--1021.

\bibitem[\protect\citeauthoryear{Lopes, Melo, and
  Montesano}{2009}]{lopes2009active}
Lopes, M.; Melo, F.; and Montesano, L.
\newblock 2009.
\newblock Active learning for reward estimation in inverse reinforcement
  learning.
\newblock In {\em Joint European Conference on Machine Learning and Knowledge
  Discovery in Databases}.

\bibitem[\protect\citeauthoryear{Michini and How}{2012a}]{michini2012bayesian}
Michini, B., and How, J.~P.
\newblock 2012a.
\newblock Bayesian nonparametric inverse reinforcement learning.
\newblock In {\em Joint European Conference on Machine Learning and Knowledge
  Discovery in Databases}.

\bibitem[\protect\citeauthoryear{Michini and How}{2012b}]{michini2012improving}
Michini, B., and How, J.~P.
\newblock 2012b.
\newblock Improving the efficiency of bayesian inverse reinforcement learning.
\newblock In {\em IEEE International Conference on Robotics and Automation},
  3651--3656.

\bibitem[\protect\citeauthoryear{Ng and Russell}{2000}]{ng2000algorithms}
Ng, A.~Y., and Russell, S.~J.
\newblock 2000.
\newblock Algorithms for inverse reinforcement learning.
\newblock In {\em Proceedings of the International Conference on Machine
  Learning},  663--670.

\bibitem[\protect\citeauthoryear{Ogryczak and
  Ruszczy{\'n}ski}{1999}]{ogryczak1999stochastic}
Ogryczak, W., and Ruszczy{\'n}ski, A.
\newblock 1999.
\newblock From stochastic dominance to mean-risk models: Semideviations as risk
  measures.
\newblock {\em European Journal of Operational Research} 116(1):33--50.

\bibitem[\protect\citeauthoryear{Pirotta and
  Restelli}{2016}]{pirotta2016inverse}
Pirotta, M., and Restelli, M.
\newblock 2016.
\newblock Inverse reinforcement learning through policy gradient minimization.
\newblock In {\em Proceedings of the AAAI Conference on Artificial
  Intelligence},  1993--1999.

\bibitem[\protect\citeauthoryear{Ramachandran and
  Amir}{2007}]{ramachandran2007bayesian}
Ramachandran, D., and Amir, E.
\newblock 2007.
\newblock Bayesian inverse reinforcement learning.
\newblock In {\em Proceedings of the 20th International Joint Conference on
  Artifical intelligence},  2586--2591.

\bibitem[\protect\citeauthoryear{Rockafellar and
  Uryasev}{2000}]{rockafellar2000optimization}
Rockafellar, R.~T., and Uryasev, S.
\newblock 2000.
\newblock Optimization of conditional value-at-risk.
\newblock {\em Journal of risk} 2:21--42.

\bibitem[\protect\citeauthoryear{Rothkopf and
  Ballard}{2013}]{rothkopf2013modular}
Rothkopf, C.~A., and Ballard, D.~H.
\newblock 2013.
\newblock Modular inverse reinforcement learning for visuomotor behavior.
\newblock {\em Biological cybernetics} 107(4):477--490.

\bibitem[\protect\citeauthoryear{Santara \bgroup et al\mbox.\egroup
  }{2017}]{santara2017rail}
Santara, A.; Naik, A.; Ravindran, B.; Das, D.; Mudigere, D.; Avancha, S.; and
  Kaul, B.
\newblock 2017.
\newblock Rail: Risk-averse imitation learning.
\newblock {\em arXiv preprint arXiv:1707.06658}.

\bibitem[\protect\citeauthoryear{Syed and Schapire}{2008}]{syed2008game}
Syed, U., and Schapire, R.~E.
\newblock 2008.
\newblock A game-theoretic approach to apprenticeship learning.
\newblock In {\em Advances in neural information processing systems},
  1449--1456.

\bibitem[\protect\citeauthoryear{Tamar, Glassner, and
  Mannor}{2015}]{tamar2015optimizing}
Tamar, A.; Glassner, Y.; and Mannor, S.
\newblock 2015.
\newblock Optimizing the cvar via sampling.
\newblock In {\em Proceedings of the Twenty-Ninth AAAI Conference on Artificial
  Intelligence},  2993--2999.

\bibitem[\protect\citeauthoryear{Thomas \bgroup et al\mbox.\egroup
  }{2017}]{thomas2017ensuring}
Thomas, P.~S.; da~Silva, B.~C.; Barto, A.~G.; and Brunskill, E.
\newblock 2017.
\newblock On ensuring that intelligent machines are well-behaved.
\newblock {\em arXiv preprint arXiv:1708.05448}.

\bibitem[\protect\citeauthoryear{Thomas, Theocharous, and
  Ghavamzadeh}{2015a}]{thomas2015improvement}
Thomas, P.; Theocharous, G.; and Ghavamzadeh, M.
\newblock 2015a.
\newblock High confidence policy improvement.
\newblock In {\em Proceedings of the 32nd International Conference on Machine
  Learning},  2380--2388.

\bibitem[\protect\citeauthoryear{Thomas, Theocharous, and
  Ghavamzadeh}{2015b}]{thomas2015high}
Thomas, P.~S.; Theocharous, G.; and Ghavamzadeh, M.
\newblock 2015b.
\newblock High-confidence off-policy evaluation.
\newblock In {\em Proceedings of the AAAI Conference on Artificial
  Intelligence},  3000--3006.

\bibitem[\protect\citeauthoryear{Weisstein}{2017}]{weissteinball}
Weisstein, E.
\newblock 2017.
\newblock Ball point picking.
\newblock {\em From MathWorld--A Wolfram Web Resource. http://mathworld.
  wolfram. com/BallPointPicking. html}.

\bibitem[\protect\citeauthoryear{Wulfmeier, Ondruska, and
  Posner}{2015}]{wulfmeier2015maximum}
Wulfmeier, M.; Ondruska, P.; and Posner, I.
\newblock 2015.
\newblock Maximum entropy deep inverse reinforcement learning.
\newblock {\em arXiv preprint arXiv:1507.04888}.

\bibitem[\protect\citeauthoryear{Zheng, Liu, and Ni}{2014}]{zheng2014robust}
Zheng, J.; Liu, S.; and Ni, L.~M.
\newblock 2014.
\newblock Robust bayesian inverse reinforcement learning with sparse behavior
  noise.
\newblock In {\em Proceedings of the AAAI Conference on Artificial
  Intelligence},  2198--2205.

\bibitem[\protect\citeauthoryear{Ziebart \bgroup et al\mbox.\egroup
  }{2008}]{ziebart2008maximum}
Ziebart, B.~D.; Maas, A.~L.; Bagnell, J.~A.; and Dey, A.~K.
\newblock 2008.
\newblock Maximum entropy inverse reinforcement learning.
\newblock In {\em Proceedings of the 23rd AAAI Conference on Artificial
  Intelligence},  1433--1438.

\end{thebibliography}

\appendix
\section{Appendix}

\subsection{Uniform sampling from L1-unit ball}
We derive an algorithm that correctly samples uniformly form the L1-norm unit ball. Our method is a special case of the result by Barthe et al. \cite{barthe2005probabilistic} as detailed in Weisstein \cite{weissteinball}. The general result states that if we wish to sample an element from the L-p ball in d-dimensional space, then we should pick $X_1, \ldots, X_d$ independently from the pdf
\begin{equation}
P_p(x) = \frac{\exp(-|x|^p)}{2 \Gamma(1 + p^{-1})}
\end{equation}
where $p$ is the desired $p$-norm and $\Gamma$ is the gamma function. Then we draw $Y$ from an exponential distribution with mean 1 and our resulting sample from the $L_p$ norm ball is 
\begin{equation}
\frac{(X_1,\ldots, X_n)}{(Y + \sum_{i=1}^n |X_i|^p)^{1/p}}
\end{equation}

We wish to sample from the L1-norm boundary, i.e. where the L1-norm is equal to 1. Thus we have $p=1$ and $Y=0$ above. This means that we need to sample $d$ numbers independently from the following pdf
\begin{eqnarray}
P_1(x) = \frac{\exp(-|x|)}{2 \Gamma(2)} = \frac{\exp(-|x|)}{2}.
\end{eqnarray}

We can sample from this distribution using the inverse CDF sampling method (c.f. Bishop \cite{bishop2006pattern}). 
To draw samples from this distribution we must compute the inverse of the indefinite integral
\begin{equation}
z = h(x) = \int_{-\infty}^x \frac{\exp(-|\hat{x}|)}{2} d\hat{x}
\end{equation}

Note that the desired distribution, $P_1(x)$, is a peaked distribution centered at zero, so half of the probability mass will be less than zero and half will be greater than zero. We can thus break-up our inverse of the CDF into two cases.

Case 1: If our random uniform sample $z \in [0,1/2]$, then our resulting $x$ should be non-positive. In this case we can write $P_1(x)$ as
\begin{equation}
P^-_1(x) = \frac{\exp(x)}{2}
\end{equation}
We can now easily solve for $f(z) = h^{-1}(z)$ where
\begin{equation}
z = h(x) = \frac{1}{2}\int_{-\infty}^x \exp(\hat{x}) d\hat{x}.
\end{equation}
Solving the integral and inverting gives
\begin{equation}
x = \ln(2z).
\end{equation}

Case 2: $z \in [1/2,1]$. In this case, $x$, our resulting sample from $P_1(x)$ should be non-negative. Thus, we can write $P_1(x)$ as
\begin{equation}
P^+_1(x) = \frac{\exp(-x)}{2}
\end{equation}
We can now solve for $f(z) = h^{-1}(z)$ where this time
\begin{eqnarray}
z = h(x) &=& \frac{1}{2}\int_{-\infty}^x \exp(-\hat{x}) \\
&=& \frac{1}{2} + \frac{1}{2}\int_{0}^x \exp(-\hat{x}) d\hat{x}.
\end{eqnarray}
Solving the and inverting gives
\begin{equation}
x = -\ln(2-2z).
\end{equation}

In summary, to sample from $P_1(x) = \frac{\exp(-|x|)}{2}$ we first draw $z \sim [0,1]$. Then we return
\begin{equation}
x = \begin{cases}
\ln(2z),& \quad  \text{for } z < 1/2 \\
-\ln(2-2z),& \quad \text{otherwise}
\end{cases}
\end{equation}

Using $d$ samples from $P_1(x)$ and then normalize the resulting sample gives us a way to uniformly sample the L1-norm unit sphere. This method summarized in Algorithm~\ref{alg:L1SphereSampling} for uniformly sampling from the L1 unit sphere. 

\begin{algorithm}
\caption{L1-Norm Unit Ball Sampling in $\mathbb{R}^d$}\label{alg:L1SphereSampling}
\begin{algorithmic}[1]
\State \textbf{input:} $d$ \Comment number of dimensions 
\For{$i = 1:d$} 
\State $z \sim U(0,1)$
\If{$z \leq 0.5$}
\State $X_i = \ln(2z)$
\Else
\State  $X_i = -\ln(2 - 2z)$
\EndIf
\EndFor
\State $\mathbf{X} \gets (X_1,\ldots, X_d)/\sum_{i=1}^d |X_i|$
\State\Return $\mathbf{X}$
\end{algorithmic}
\end{algorithm}

\subsection{MCMC implementation details}
Our MCMC implementation of BIRL ensures that each proposal step remains on the L1-norm unit ball. We use Algorithm~\ref{alg:L1MCMCWalk} to generate a proposal by taking a small step along each pair of axis while staying on the L1-norm unit ball. For each pair of axis we use Algorithm~\ref{alg:L1ManifoldStep} to step along the manifold defined by the two axis.

In all of our grid world experiments we use stepSize = 0.01 for the L1-Norm Unit Ball Walk described in Algorithm~\ref{alg:L1MCMCWalk}. We run MCMC for 10000 steps using a burn-in of 100 samples and only using every 20th sample to avoid autocorrelation effects.

We found that the BIRL likelihood can be sensitive to data imbalance if the demonstrations contain some state-action pairs much more frequently than others. To ameliorate this problem, we remove duplicate state-action pairs from the demonstrations.

\begin{algorithm}[t]
\caption{L1-Norm Unit Ball Walk}\label{alg:L1MCMCWalk}
\begin{algorithmic}[1]
\State \textbf{input:} $w \in \mathbb{R}^d$, stepSize \Comment initial weight vector
\For{ each pair of dimensions $(i,j):$ $i,j=1,\ldots,d$} 
\State direction $\gets$ random('clockwise', 'counterclockwise')
\If{$w[i]$ \textbf{is not} $0$ \textbf{or} $w[j]$ \textbf{is not} $0$}
\State $w[i], w[j] \gets$ L1ManifoldStep($w[i]$, $w[j]$, direction, $\eta$)
\EndIf
\EndFor
\State\Return $w$
\end{algorithmic}
\end{algorithm}

\begin{algorithm*}[t]
\caption{L1ManifoldStep}\label{alg:L1ManifoldStep}
\begin{algorithmic}[1]
\State \textbf{input:} $w1, w2 \in \mathbb{R}$, direction$\in\{$clockwise, counterclockwise$\}$, stepSize $\in \mathbb{R}$
\State slack = w1 + w2
\State clockwisePos = ["+ +","+ -","- -","- +"]
\State clockwiseDir = [+1,-1,+1,-1]
\State counterclockwisePos =  ["+ +","- +","- -","+ -"]
\State counterclockwiseDir = [-1,+1,-1,+1]
\State sign1 = (w1 $\geq$ 0) \Comment find starting quadrant
\State sign2 = (w2 $\geq$ 0)
\If{sign1 \textbf{and} sign2}
\State cyclePos = "+ +"
\ElsIf{sign1 \textbf{and not} sign2}
\State cyclePos = "+ -";
\ElsIf{\textbf{not} sign1 \textbf{and} sign2}
\State cyclePos = "- +"
\Else 
\State cyclePos = "- -"
\EndIf    
\If{direction \textbf{is} "clockwise"} \Comment find direction to change magnitude of w1
\State cycleIndx = clockwisePos.indexOf(cyclePos)
\Else
\State cycleIndx = counterclockwisePos.indexOf(cyclePos)
\EndIf
\State stepRemaining = stepSize
\While{stepRemaining $>$ 0} \Comment step along 1-D manifold of L1-unit ball in 2-D
\If{direction \textbf{is} "clockwise"}
\State cycleDir = clockwiseDir[cycleIndx]
\Else
\State cycleDir = counterclockwiseDir[cycleIndx]
\EndIf
\State maxStep = stepRemaining
\If{cycleDir \textbf{is} 1 \textbf{and} ($|w1|$ + cycleDir * stepRemaining) $>$ slack}
\State maxStep = slack - $|w1|$
\State cycleIndx = mod(cycleIndx + 1, 4) 
\ElsIf{cycleDir \textbf{is} -1 \textbf{and} ($|w1|$ + cycleDir * stepRemaining) $<$ 0}
\State maxStep = $|w1|$
\State cycleIndx = mod(cycleIndx + 1, 4) 
\EndIf
\State w1 = $|w1|$ + cycleDir * maxStep
\State w2 = $|w2|$ - cycleDir * maxStep
\State stepRemaining = stepRemaining - maxStep
\EndWhile
\If{randDir \textbf{is} "clockwise"} \Comment determine correct signs based on final quadrant
\State cyclePos = clockwisePos[cycleIndx]
\Else
\State cyclePos = counterclockwisePos[cycleIndx]
\EndIf
\If{cyclePos \textbf{is} "- +"}
\State w1 = -w1 
\ElsIf{cyclePos \textbf{is} "+ -"}
\State w2 = -w2
\ElsIf{cyclePos \textbf{is} "- -"}
\State w1 = -w1
\State w2 = -w2
\EndIf
\State\Return $w1$, $w2$
\end{algorithmic}
\end{algorithm*}

\subsection{Sensitivity to evaluation policy} \label{appendix:evalPolicySensitivity}
In this section we investigate how the bound on VaR is affected by the choice of evaluation policy. We ran an experiment where we varied the optimality of the evaluation policy. As in our previous experiments, used a 9x9 grid world and we generated 200 MDPs with random features and feature weights for evaluation. The evaluation policy was chosen by taking the optimal policy obtained from the ground truth reward and selecting $X$ states at random without replacement and changing the policy at those $X$ states so that it takes a non-optimal action. All demonstrations were optimal so we computed the VaR bounds using $c=100$.

\begin{figure}[t]
\centering
\subfigure[Accuracy after one demonstration]
{\includegraphics[scale=0.25]
{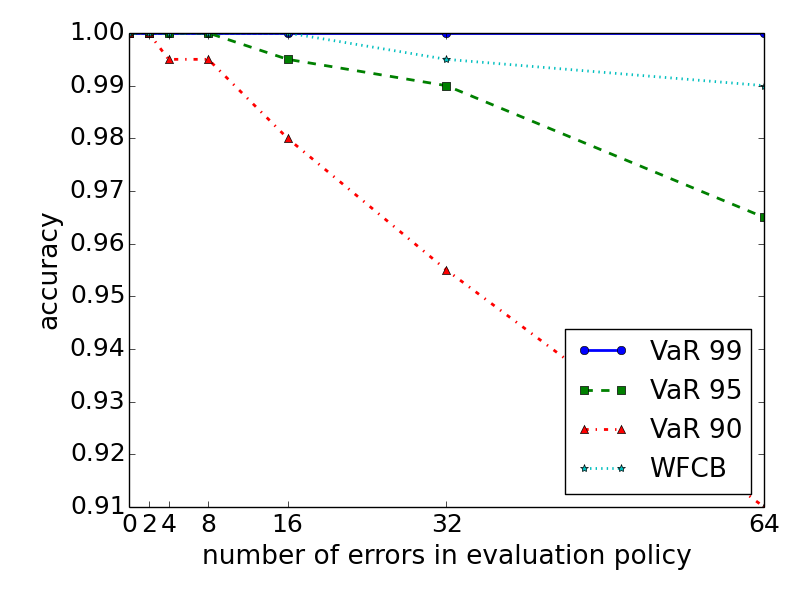}
\label{subfig:eval_acc1}}
\subfigure[Average bound error after one demonstrations]
{\includegraphics[scale=0.25]
{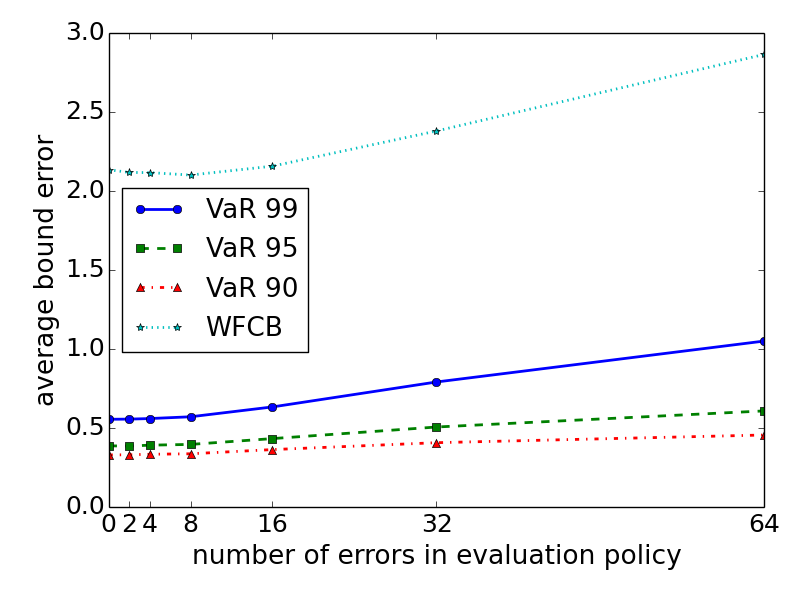}
\label{subfig:eval_error1}}
\caption{Sensitivity for bounding the performance of a range of evaluation policies given 1 optimal demonstration. Results are averaged over 200 grid navigation task with no terminal states. Accuracy and average error for WFCB bounds versus bounds on the 0.99-, 0.95-, and 0.90-VaR.}
\label{fig:evalPolicies_1demo}
\end{figure}

\begin{figure}[t]
\centering
\subfigure[Accuracy after nine demonstrations]
{\includegraphics[scale=0.25]
{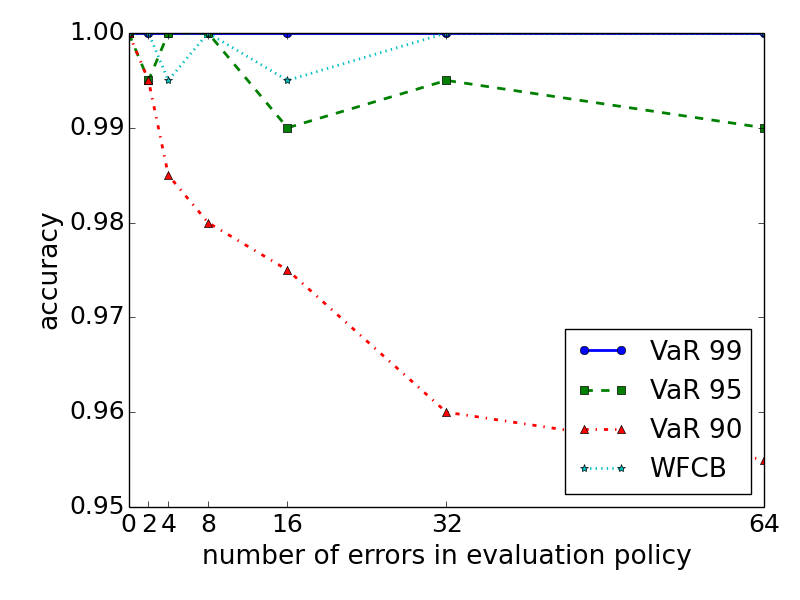}
\label{subfig:eval_acc9}}
\subfigure[Average bound error after nine demonstrations]
{\includegraphics[scale=0.25]
{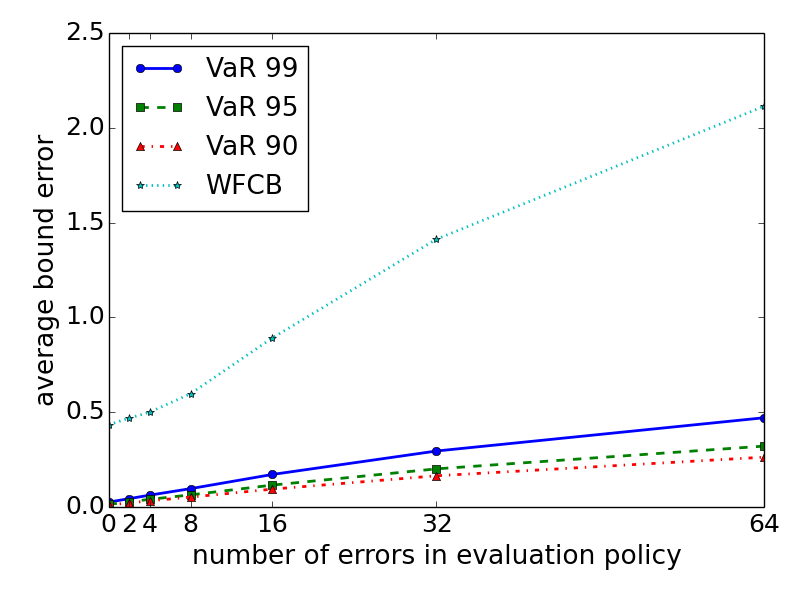}
\label{subfig:eval_error9}}
\caption{Sensitivity for bounding the performance of a range of evaluation policies given 9 optimal demonstrations. Results are averaged over 200 grid navigation task with no terminal states. Accuracy and average error for WFCB bounds versus bounds on the 0.99-, 0.95-, and 0.90-VaR.}
\label{fig:evalPolicies_9demos}
\end{figure}

The results for $X=0,2,4,8,16,32,64$ after one demonstration are shown in Figure~\ref{fig:evalPolicies_1demo} and the results after nine demonstrations are shown in Figure~\ref{fig:evalPolicies_9demos}. When the evaluation matches the optimal policy under the true reward ($X = 0$) all bound methods always gave true upper bounds on the EVD. When only one demonstration is given, there is a large bound error for all methods, with WFCB giving an error bound 4 times higher than the worst VaR bound error. As $X$ is increased, the evaluation policy becomes more dissimilar to the optimal policy. This results in a drop in accuracy for all bounds except for the extremely conservative 0.99-VaR bound. 
When 9 demonstrations are given, the VaR bounds are much tighter with almost zero error for evaluation bounds close to optimal. The accuracy tends to drop as the number of errors increases, but still remains above 95\% even for a policy that differs from the optimal policy in over 75\% of the states.

\subsection{Using the Projection algorithm to obtain evaluation policy}
Abbeel and Ng \shortcite{abbeel2004apprenticeship} give sample efficiency bounds for the number of demonstrations required to get within $\epsilon$ of an experts performance. We summarize their result as the following theorem.
\begin{theorem} \cite{abbeel2004apprenticeship}
To obtain a policy $\hat{\pi}$ such that with probability $(1-\delta)$
\begin{equation}
\epsilon \geq |V^{\hat{\pi}}(R^*) - V^{\pi^*}(R^*)|
\end{equation} 
it suffices to have 
\begin{equation}
m \geq \frac{2k}{(\epsilon(1-\gamma))^2}\log \frac{2k}{\delta}.
\end{equation}
\end{theorem}

Often we are simply given a fixed set of demonstrations and wish to know how far from optimal performance our learn policy is. Using a similar proof we obtained the following corollary for a fixed number of demonstrations and a desired confidence level, $\delta$.
\begin{corollary}\label{cor:abbeel_fixed_num_demos}
Given a confidence level $\delta$, and $m$ demonstrations, with probability $(1-\delta)$ we have that $|V^{\pi^*}(R^*) - V^{\hat{\pi}}(R^*)| \leq \epsilon$, where 
\begin{equation}
\epsilon \leq \frac{1}{1-\gamma}\sqrt{\frac{2k}{m}\log{\frac{2k}{\delta}}}
\end{equation}
where $k$ is the number of features and $\gamma$ is the discount factor of the underlying MDP.
\end{corollary}

Syed and Schapire \shortcite{syed2008game} proved an even tighter bound that holds for the Projection algorithm as well as their Multiplicative Weights Algorithm. We summarize their result as the following theorem.
\begin{theorem} \cite{syed2008game}
To obtain a policy $\hat{\pi}$ such that with probability $(1-\delta)$
\begin{equation}
\epsilon \geq |V^{\hat{\pi}}(R^*) - V^{\pi^*}(R^*)|
\end{equation} 
it suffices to have 
\begin{equation}
m \geq \frac{2}{(\frac{\epsilon}{3}(1-\gamma))^2}\log \frac{2k}{\delta}.
\end{equation}
\end{theorem}

We inverted this inequality to obtain the following corollary which is used as a benchmark in our paper.

\begin{corollary}\label{cor:syed_fixed_num_demos}
Given a confidence level $\delta$, and $m$ demonstrations, with probability $(1-\delta)$ we have that $|V^{\pi^*}(R^*) - V^{\hat{\pi}}(R^*)| \leq \epsilon$, where 
\begin{equation}
\epsilon \leq \frac{3}{1-\gamma}\sqrt{\frac{2}{m}\log{\frac{2k}{\delta}}}
\end{equation}
where $k$ is the number of features and $\gamma$ is the discount factor of the underlying MDP.
\end{corollary}

\end{document}